# Infrastructure for the representation and electronic exchange of design knowledge

Laurent BUZON*, Abdelaziz BOURAS, Yacine OUZROUT

IUT Lumière, Université Lyon2

Laboratory PRISMA/CERRAL

160 Bd de l'Université 69676 Bron cedex - France

Email : first name.last name@univ-lyon2. Fr

Phone : (+33) 4 78 77 26 70

***Purpose of the submission*** : This paper develops the concept of knowledge and its exchange using Semantic Web technologies. It points out that knowledge is more than information because it embodies the meaning, that is to say semantic and context. These characteristics will influence our approach to represent and to treat the knowledge. In order to be adopted, the developed system needs to be simple and to use standards. The goal of the paper is to find standards to model knowledge and exchange it with an other person. Therefore, we propose to model knowledge using UML models to show a graphical representation and to exchange it with XML to ensure the portability at low cost. We introduce the concept of ontology for organizing knowledge and for facilitating the knowledge exchange. Proposals have been tested by implementing an application on the design knowledge of a pen.

***Subject/theme of the journal :***

- Product data management and standards
- Advanced technologies in e-business development
- Electronic Knowledge Management

# Infrastructure for the representation and electronic exchange of design knowledge

Laurent BUZON*, Abdelaziz BOURAS and Yacine OUZROUT

Université Lumière Lyon 2, IUT Lumière

CERRAL – PRISMA Laboratory

160, Bd de l'Université 69676 Bron cedex - France

E-mail : firstname.lastname@univ-lyon2.fr

Phone : (+33) 4 78 77 26 70

# Infrastructure for the representation and electronic exchange of design knowledge

**Abstract :**

With the growing complexity of information, organizations consider more and more collaborative aspects for completely and correctly exchanging information among different systems. These organizations involve many specialists from different domains, with wide-ranging knowledge. Consequently, knowledge is now regarded as a strategic asset which must be managed, with an increasing need for representation formalisms and deployment of tools for knowledge exchange. This paper proposes a contribution to this area that privileges the user point of view. It is based on standards and simple formalisms to ensure the user adoption and the portability of knowledge exchange. UML formalism as well as XML technology were retained for their skills to meet the interoperability needs and their capacity to be treated by users and computers. An application on design knowledge of a pen, resulting from a CYGMA "knowledge book", was carried out in order to illustrate and to validate our choices.

# 1. Introduction

A major issue in collaborative design is the creation and maintenance of a suitable representation for design knowledge that will be shared by many designers and users. Precise and unambiguous capture of the meaning of concepts within a given system is becoming apparent. Many works already dealt with these knowledge topics and highlights the need to use specific computer interfaces to facilitate the treatment and the comprehension of the knowledge by the users. These works also prescribe the use of Web technologies to allow inter-working between different systems (Dieng-Kuntz, Corby, Gandon, Giboin, Golebiowska, Matta & Ribiere, 2000). Internet technologies have been exploited since 1994 for the construction of corporate memory by Huynh (Huynh, Popkin, & Stecker, 1994) to promote the share and the exchange of knowledge. Metadata and ontologies were used to provide a context to locate and to understand available information. One could currently find these concerns in many projects, as OntoWeb[1], OnToKnowledge[2], Acacia[3].

It is from the user point of view that we directed ourselves towards the use of automatically serialized and easily comprehensible formalisms. The present work's goal is to define tools for knowledge representation and transfer within a distributed environment, and to make knowledge understandable and accessible for a design team.

The exchanged knowledge must meet the following needs:
- representing different levels of abstraction,
- being comprehensible through an easily understanding formalism,
- being created and transferred with standard data-processing tools
- being generic, reusable and easy to extend.

The use of standards as UML (Unified Modelling Language), the OMG (Object Management Group) standard for the object oriented modelling, reasonably answers these criteria. UML have capacities to model objects, processes, ontologies (Cranfield, 1999). Moreover, the UML serialization language (XMI) is a XML extension which allows the data exchange between different systems and thus ensures the portability of knowledge.

---

[1] http://www.ontoweb.org/

[2] http://www.ontoknowledge.org/

[3] http://www-sop.inria.fr/acacia/

The work described in this paper is illustrated by a case study on design knowledge of an "advertising pen" drawn from a knowledge book CYGMA[4] (Serrafero, Vargas, & Renson, 1999), where engineering knowledge have been organized (ontology), modelled (UML) and exchanged (XMI) via Internet.

## 2. The methodology

We consider that information is the support of the knowledge transfer between people or systems. Thus, knowledge is specific to each individual and it is the result of the information interpretation in its reference frame. If we follow J-L. Ermine (2000) point of view, this reference frame includes the semantics which the individual binds to this information and the context in which information is integrated. In concrete terms, only informational traces of this knowledge can be observed.

(Figure 1)

Hence, we use a methodology based on six steps, as we can see in the first figure. The process is illustrated with the design knowledge of a pen for which we have already studied the first tow step and selected the CYGMA knowledge capture methodology, to make a knowledge book. This knowledge book is intended to include knowledge concepts such as component structure, features, parameters, constraints, requirements, and more.

## 3. The knowledge representation

Cognitive modelling consists in identifying the interesting characteristics of knowledge. The abstract character of a model facilitates the understanding of the studied system. « *It reduces the complexity of the studied system, makes it possible to simulate it, to represent it and to reproduce its behaviours* "(Zacklad & Grundstein, 2001).

The knowledge engineering research has developed powered and specific languages of knowledge representation as KIF (Knowledge Interchange Format)[5] and more generic and standard languages to represent knowledge.

---

[4] CYGMA : CYcle de vie et Gestion des Métiers et des Applications (www.cegos-kadetech.fr): Methodology of knowledge management developed by Kade-tech/KAD-KAM International.

[5] http://logic.stanford.edu/kif/kif.html (Stanford univesity KIF website)

Standard languages have been chosen in this work to:

- store and organize knowledge around the object concept,
- facilitate the knowledge modelling and its treatment,
- provide low inferential services intended to manage the knowledge base or to create new knowledge

In that way, a first solution for the design knowledge representation is the ISO STEP standard. STEP proposes to provide all the resources and methods allowing to describe the whole of the product data resulting from the process of design/manufacturing/maintenance independently of a particular information processing system and to translate from a human understanding representation of product to a computer representation. STEP includes the information model specification language and its graphical notation. However, STEP does not completely feed our needs because it concentrates on the product point of view and does not model the organizational point of view (Arnold & Podehl, 1998). Moreover, in the knowledge exchange phase, we are more interested in the structure of the data than the data themselves.

The second solution is UML, which allows a representation of various knowledge point of views thanks to the use of its 9 diagrams. UML describes the modelling elements (concepts conveyed and handled by the language) and the element semantics (their definition and the direction of their use). It also makes it possible to classify the various concepts of the language (according to their level of abstraction or their applicability) and thus exposes clearly its structure.

The formal aspect of UML notation limits ambiguities and misunderstandings. *It represents a good medium between mathematical language and natural language, not too complex but sufficiently rigorous, because it's based on a meta model* (Muller, 1997).

UML proposes diagrams to describe the studied system :

- **Static diagrams** represent the properties and the relations between classes and objects (derivatives of the object model OMT). The static diagrams represent the knowledge deprived of temporal aspect. For example, ontology is represented by a class diagram.

We introduced below the example of the Lead_Protection network, which describes the relations between the concepts of this field. This diagram presents the elements that compose a lead protection.

(Figure 2)

- **Dynamic diagrams** represent temporal aspects and scenarios with sequence diagram, activity diagram, collaboration diagram, and statechart diagram.

(Figure 3 and 4)

These diagrams present knowledge that help us to understand the ink leak effect. These models explain the phenomenon with different point of view. The state chart diagram illustrates a first point of view which concentrates on the evolution of the pen state and the collaboration diagram presents a second point of view which highlights the process itself and the relations between objects.

UML Diagrams are completed by a declaratory language which describes constraints or rules. This formal language OCL (Object Constraint Language), is simple and has an elementary grammar which can be interpreted by data-processing tools. It represents a medium, between a natural language and a mathematical language. Thus OCL makes it possible to limit ambiguities, while remaining accessible.

OCL describes model invariants like pre and post-conditions for an operation, navigation expressions and boolean expressions, etc. The interior_diameter expression in OCL is given here as an example:

> **context** interior_diameter **inv** :
>
> interior_diameter = external_tip_diameter+2*(cone_length *SIN(cone_angle))

A part of the correspondence between the CYGMA typology of knowledge (Serrafero, Vargas, & Renson, 1999) and the used diagrams is found in the following table:

(Table 1)

The UML contributions for the knowledge representation are numerous:

- The knowledge expressed with UML, is directly understandable by the user (thanks to the graphical representation) and by the computer (via XMI associated API defined by the OMG),
- Knowledge can evolve easily thank to object modelling,
- New knowledge can be derived from UML models thanks to the reasoning on their contents. Particularly, the use of OCL allows to deduce constraints and rules on a diagram (Cranefiel, 2001),
- It is general, it concentrates on a whole of concepts, and also offers mechanisms of extension,
- Several tools allow to use UML (Magicdraw UML, Rational Rose,…).

However, as we can see in the Table 1, UML cannot describe each knowledge type and in order to make complete knowledge card, the diagrams are enhanced by user description (texts, references and schemas).

Once knowledge is described, our objective is to exchange it, to share it and make it available to the other team members. Web Technologies are a solution to deal with the interworking need between system and the user adoption

## 4. The Knowledge Exchange

The notion of "Semantic Web" provides enhanced information access bases on the exploitation of machine-processing meta-data to automatically carry out varied tasks (Berners-Lee, Hendler, & Lassila, 2001). This approach lays on the existence of structured or semi-structured data on the Web, represented in a formalism, authorizing automated treatments. Such treatments must support the integration of data resulting from multiple and heterogeneous sources and make possible their use in various applications (Staab, 2003).

The basic language of the semantic Web is the structured format XML[6] (eXtensible Language Mark-up) which encodes the contents of document according to a corresponding schema. In our application, knowledge is exchanged in the form of structured cards (the structure elements are defined by a XML Schema) according to their nature (vocabulary, history, expertises, process).

(Figure 5)

XML also allows us to dynamically build knowledge cards with data of various sources: CAD models (technical diagram), drawings, reports, schemas etc. XML plays the role of an integrating tool to give an uniform vision and access, and propose treatment tools on information (Scykman, Senfaute & Sriram , 1999).

Moreover, XMI is the UML serialization language. It transforms a MOF description into document schemas and document type definitions. UML model can then be declined in the form of a XML document which respects these definitions.

In order to make sure that the whole of the users of the system makes the same semantic distinctions and uses the same terms with the same definition, the Semantic Web uses the contribution of ontologies. Ontologies are metadata schemas, providing a controlled vocabulary of terms, each with an explicitly defined and processing semantics. An ontology is usually described with a graph that presents the relation between concepts and a dictionary that shows the definition of these concepts. By defining shared and common domain concepts, they help both people and machines to communicate (Bachimond, 2002). Ontology is used as unifying framework for different points of view. It clarifies a terminology and a conceptualization shared by a given community in organization (Uschold, King, Moralee & Zorgios, 1998). In concrete terms, ontology enables that the "cap" of the pen should not be confused with the "cap" of the clothing domain.

---

[6] www.w3.org/XML/ (XML Specification).

By using XML, we could resort to several different languages[7] to describe ontology. We are interested in this work in a simple ontology, to provide a formal description and terminology for design that can be shared by all the engineers involved in the project, to achieve a high level collaboration. Of course, a more complete ontology may explicitly define the domain context, the objects' behaviours, etc.

We choose RDF(S)[8] because it provides a solution for the information description from a semantic point of view on the Web. RDF(S) is simple to handle (written in XML) and is used outside the IA community for the annotation of documents. It is however less powerful for the expression of ontology and for reasoning process but it corresponds to the need of simplicity for the knowledge edition.

RDF Schema is dedicated to the specification of schemas representing the ontological knowledge used in RDF statements, it defines tags used in RDF.

By way of illustration, we describe the property "composition" as follows:

| The property "composition" makes it possible to define a relation of strong aggregation between two resources | `<rdf: Property rdf: ID="composition">`<br>`    <rdf: subPropertyOf rdf: resource="# semantique_metier "/>`<br>`    <label xml: Lang="Fr">relation of strong aggregation between two resources</label>`<br>`</rdf: Property>` |
|---|---|

This property is then used to characterize the relations between the Web resources (documents or parts of documents XML in our case).

---

[7] Daml-oil Ontology (DARPA Agent Mark Up Language), OML (Ontology Markup Language), CKML (Conceptual Knowledge Markup Language), XOL (XML-based Ontology exchange Language), RDF (Resource Description Language) etc.

[8] HTTP:www.W3.org/RDF/ (W3C RDF recommendation).

| The introduction of the RDF document indicates the ontologies used by referring to a URL. The last URL refers to the specific RDF Schema that we developed | `<?xml version="1.0" encoding="UTF-8"?>`<br>`<rdf:RDF`<br>`xmlns:rdf=http://www.w3.org/1999/02/22-rdf-syntax-ns#`<br>`xmlns:rdfs=http://www.w3.org/2000/01/rdf-schema#`<br>`xmlns:dc=http://purl.org/DC/`<br>`xmlns:lb="http://localhost/rdfs/lbn-v1.2#">` |
|---|---|

The first part of the RDF document refers to RDF Schema of Dublin-Core[9]. This RDF Schema is a standard for the description of web document and defines the terms : title, author, creation, date, etc. to facilitate research on internet.

The second part of RDF document refers to RDF Schema "**lbn-v1.2**" that we have developed. It shows the relations between knowledge cards (each knowledge card describes a concept as a dictionary) in the Lead_Protection network.

(Figure 6)

| | |
|---|---|
| Described URL → <br><br> Relation type → <br><br> Descriptive URL → | `<!—It represents a part of the network -->`<br>`<rdf: Description about="http//localhost/Lead_protection">`<br>`  <lb: aggregation>`<br>`    <rdf: bag>`<br>`      <rdf. Li resource="http//localhost/mecanism"/>`<br>`      <rdf. Li resource="http//localhost/Cap"/>`<br>`    </rdf: bag>`<br>`  </lb: aggregation>`<br>`</rdf: Description>`<br>`<rdf: Description about="http//localhost/Cap">`<br>`  <lb: composition>`<br>`    <rdf: bag>`<br>`      <rdf. Li resource="http//localhost/Closer"/>`<br>`      <rdf. Li resource="http//localhost/clip"/>`<br>`    </rdf: bag>`<br>`  </lb: composition>`<br>`</rdf: Description>`<br>`</rdf: RDF>` |

RDF documents can refer to several RDF Schema ontologies as in this example where we refer to both Dublin Core standard and to the one we developed.

(Figure : 7)

The application architecture (Figure 8) is based on the architecture developed in Bouras, Ouzrout, & Wacquet, (2001) and uses a three tiers structure, where a clear separation between web clients, middle-tier application server and storage level is made.

From the web client point of view, two types of operations are possible : a knowledge viewing mode and a knowledge capture mode.

---

[9] HTTP:dublincore.org / (Dublin Core Metadata Initiative.)

In the capture mode, a remote designer creates a structured knowledge card as given in the figure 5, which contains knowledge elements that will be separately saved in different databases. This store knowledge can then be viewed to be used in a design process or only for a consultation purpose.

The viewing request of a pen top for example is treated by the application server which look for the related knowledge cards references (in the RDF level). In a second step, the application server recreates the needed knowledge cards and return them via the web server with an XSL sheet corresponding to the user preferences and the user system. The transmitted pen top cards describe its structure elements, the problems the user can encounter, the constraints and the rules he has to respect (as seen in figures 2-4), etc.

(Figure : 8)

## 5. Conclusion and future work

An illustration of a CYGMA "knowledge book" is presented in this paper on an example of a pen design.

It showed that UML formalism allows to represent knowledge and to formalize rules and constraints by using the OCL language. The use of XML and associated technologies appeared conclusive for the knowledge exchange and share through the Web. Once a user has modelled his knowledge on a knowledge card, he can exchange it with the design team or any user, using the viewing interface based on semantic web technologies.

In spite of some expressivity weaknesses (compared to other representation models of knowledge language such as logics of descriptions), RDF language family seems to be adequate for the semantic management of Web documents. However, the priority given to the portability and the user adoption of the system made difficult the reasoning implementation and will require, in our case, the use of extensions like OWL (Web Ontology Language)[10]. This work will take place in a more global project concerning the development of an exchange information place, using the semantic Web technologies to link Design Chains to Supply Chains.

## Acknowledgement

The authors wish to thank Patrick SERRAFERO and KAD/KAM *International* for giving their support to the project.

---

[10] http://www.w3.org/2001/sw/WebOnt/ (W3C website about OWL specification).

# Figures

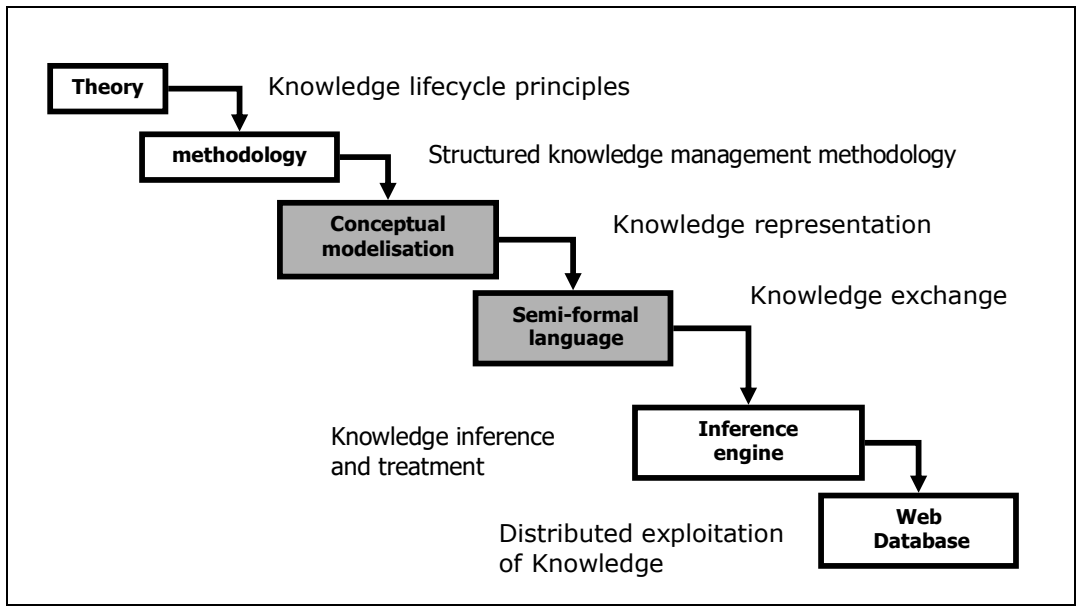

**Figure 1 :** Six steps methodology

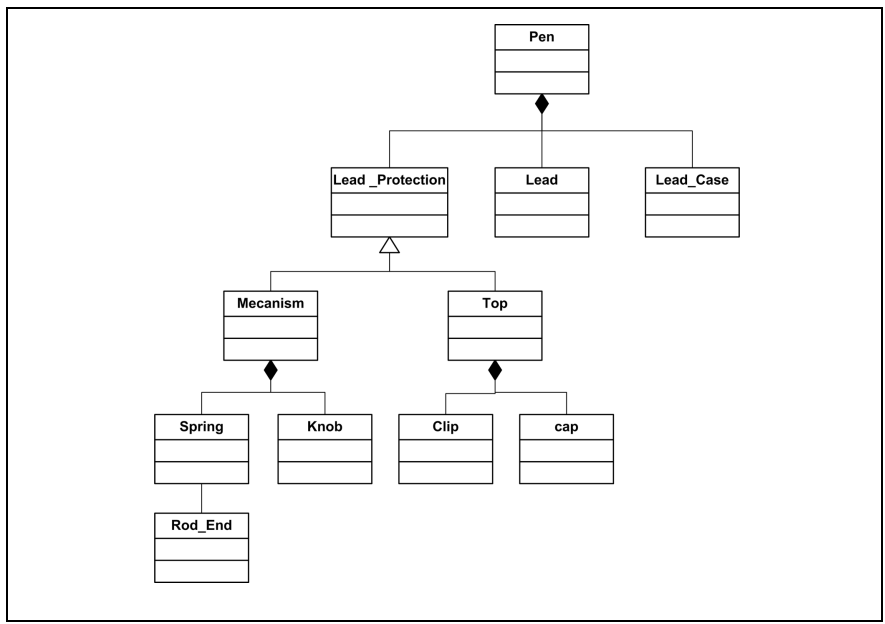

**Figure 2 :** Class diagram of lead_protection

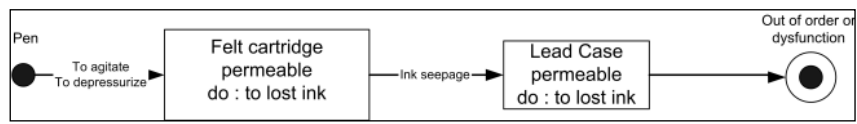

**Figure 3 :** Statechart diagram of the ink leak effect

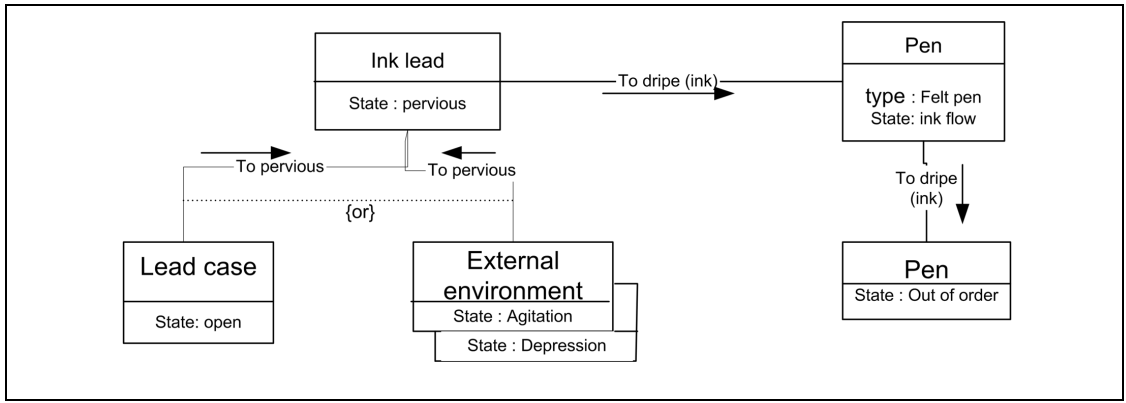

**Figure 4 :** Collaboration diagram describing the ink leak effect

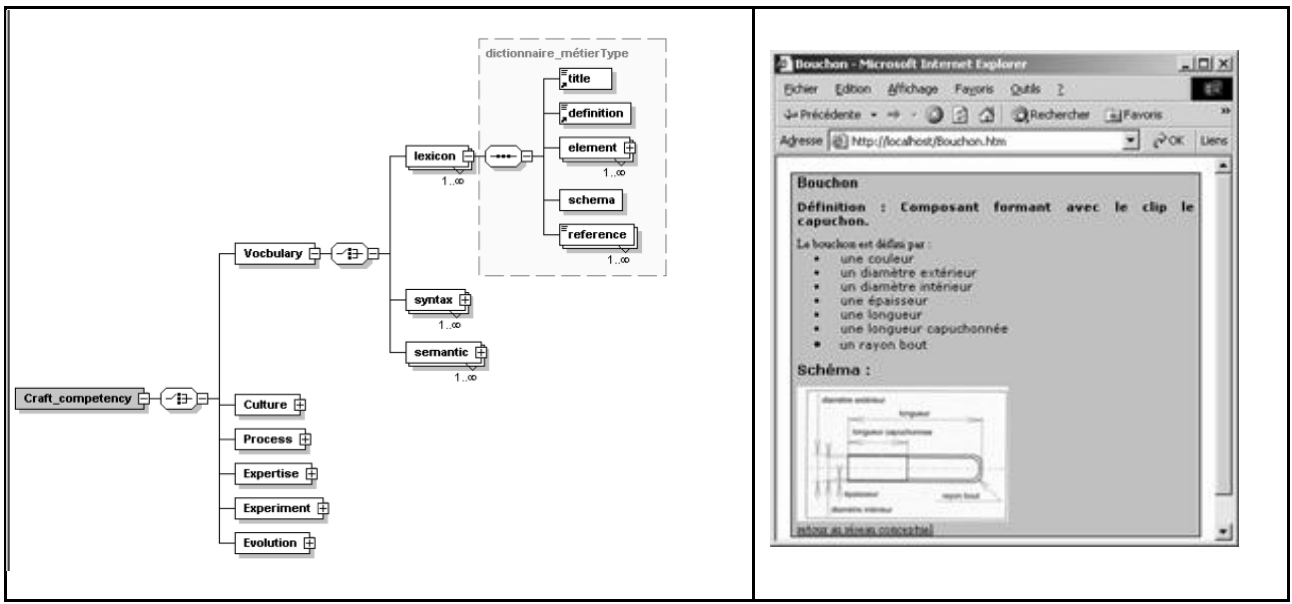

**Figure 5 :** XML schema and lexicon knowledge card

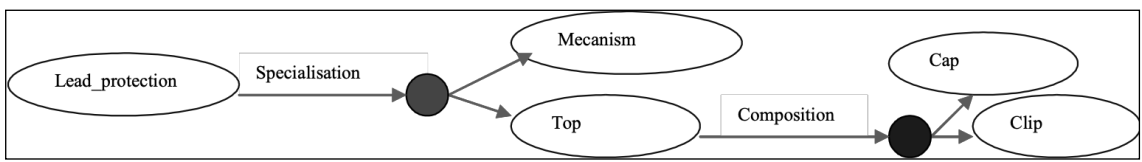

**Figure 6 :** An extract of RDF Graph describing lead_protection

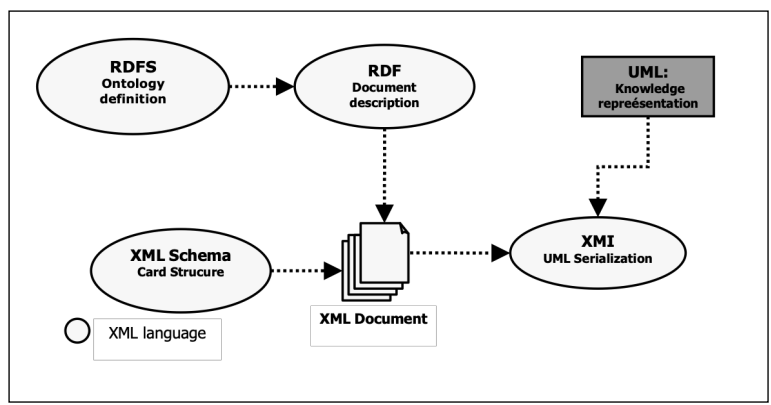

**Figure 7** : Relation between formalisms and languages

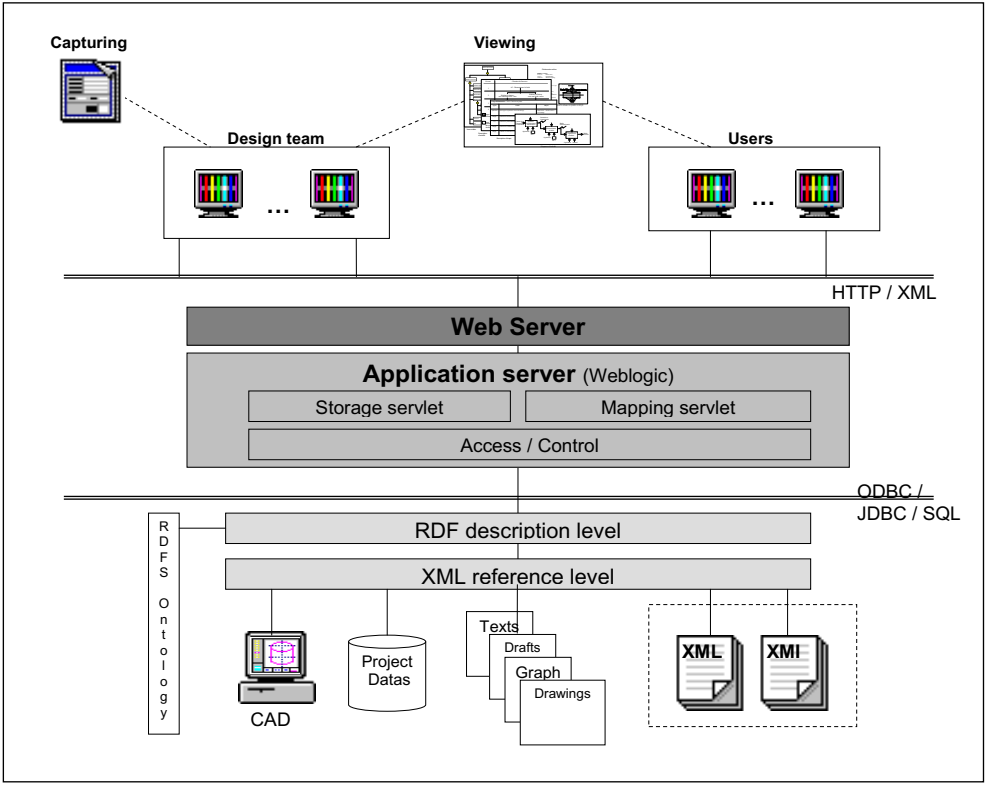

**Figure 8** : The application architecture

# Figure captions

Figure 1 : Six steps methodology

Figure 2 : Class diagram of lead_protection

Figure 3 : Statechart diagram of the ink leak effect

Figure 4 :Collaboration diagram describing the ink leak effect

Figure 5 : XML Schema and lexicon knowledge card

Figure 6 : An extract of RDF Graph describing lead_protection

Figure 7 : Relation between formalisms and languages

Figure 8 : The application architecture

# TABLES

| Type of Knowledge | | Description | Formalism |
|---|---|---|---|
| Domain Culture | History | Story | Text + schema |
| | Geography | Limit of craft knowledge | **Class Diagram** |
| | Physic | physic/chimic Phenomenon | **Statechart diagram and Collaboration Diagram** |
| Domain Process | Strategy | General cartography of the process craft | **Collaboration Diagram** |
| | Tactic | Sequence of activities | **Statechart Diagram** |
| | Diary | Description of activities | **Object Diagram and Statechart Diagram** |
| Domain Appraise | Payment | Constraint to be respected | **OCL** and Text + schema |
| | Use | Advices to be respected | **OCL** and Text + schema |
| | Freedom | Choice to be made | Text + schema |
| Domain Vocabulary | Semantics | Semantic network (Ontology) | **Class Diagram** |
| | Syntax | Gathering of sight | **Class Diagram** |
| | Lexicon | Dictionary | Text + schema |

**Table 1 :** UML diagrams and knowledge typology